\pdfoutput=1

\documentclass[11pt]{article}

\usepackage[final, nohyperref]{acl}
\usepackage{times}
\usepackage{latexsym}

\usepackage[T1]{fontenc}

\usepackage[utf8]{inputenc}

\usepackage{microtype}

\usepackage{inconsolata}

\usepackage{graphicx}

\usepackage{array}
\usepackage{amsfonts}
\usepackage{amsmath}
\usepackage{booktabs}
\usepackage{multirow}
\usepackage{paralist}
\usepackage[ruled,linesnumbered]{algorithm2e}
\usepackage{tablefootnote}
\usepackage{xcolor,colortbl}
\usepackage{scrextend}
\usepackage[colorlinks=true, citecolor=darkblue, linkcolor=darkblue, urlcolor=darkblue]{hyperref}
\usepackage{soul}

\newcolumntype{L}[1]{>{\raggedright\let\newline\\\arraybackslash\hspace{0pt}}m{#1}}
\newcolumntype{C}[1]{>{\centering\let\newline\\\arraybackslash\hspace{0pt}}m{#1}}
\newcolumntype{R}[1]{>{\raggedleft\let\newline\\\arraybackslash\hspace{0pt}}m{#1}}

\DeclareMathOperator*{\argmax}{\arg\max}

\newcommand{\spider}{S\textsc{pider}\xspace}
\newcommand{\spiderdk}{S\textsc{pider}-DK\xspace}
\newcommand{\spiderreal}{S\textsc{pider}-R\textsc{eal}\xspace}
\newcommand{\spidersyn}{S\textsc{pider}-S\textsc{yn}\xspace}
\newcommand{\cspider}{CS\textsc{pider}\xspace}
\newcommand{\astres}{\textsc{ASTReS}\xspace}

\SetCommentSty{mycommfont}

\definecolor{darkblue}{rgb}{0, 0, 0.5}
\definecolor{pale-green}{RGB}{152, 251, 152}
\definecolor{pale-orange}{RGB}{253, 200, 131}
\definecolor{pale-blue}{RGB}{209, 237, 242}
\newcommand{\red}[1]{\textcolor{red}{#1}}

\title{Improving Retrieval-augmented Text-to-SQL with AST-based Ranking and Schema Pruning}

\author{Zhili Shen$^{\mathbf{\dagger}}$ \quad Pavlos Vougiouklis$^{\mathbf{\dagger}}$  \quad Chenxin Diao$^{\mathbf{\dagger}}$ \quad {\bf Kaustubh Vyas}\\ \quad {\bf Yuanyi Ji} \quad {\bf Jeff Z. Pan}\\ 
Huawei Technologies\\Edinburgh RC, CSI
\\Edinburgh, United Kingdom\\\texttt{\{}{\href{mailto:zhilishen@huawei.com}{\texttt{zhilishen}}}\texttt{,} \href{mailto:pavlos.vougiouklis@huawei.com}{\texttt{pavlos.vougiouklis}}\texttt{,} \href{mailto:chenxindiao@huawei.com}{\texttt{chenxindiao}}\texttt{,} \href{mailto:kaustubh.vyas@huawei.com}{\texttt{kaustubh.vyas}}\texttt{\}@huawei.com}\\  \texttt{\{}\href{mailto:jiyuanyi@huawei.com}{\texttt{jiyuanyi}}\texttt{,}  \href{mailto:jeff.pan@huawei.com}{\texttt{jeff.pan}}\texttt{\}@huawei.com}\\}

\begin{document}
\maketitle
\begin{abstract}

We focus on Text-to-SQL semantic parsing from the perspective of retrieval-augmented generation. Motivated by challenges related to the size of commercial database schemata and the deployability of business intelligence solutions, we propose \astres that dynamically retrieves input database information and uses abstract syntax trees to select few-shot examples for in-context learning.

Furthermore, we investigate the extent to which an in-parallel semantic parser can be leveraged for generating \textit{approximated} versions of the expected SQL queries, to support our retrieval. We take this approach to the extreme---we adapt a model consisting of less than $500$M parameters, to act as an extremely efficient approximator, enhancing it with the ability to process schemata in a parallelised manner. We apply \astres to monolingual and cross-lingual benchmarks for semantic parsing, showing improvements over state-of-the-art baselines. Comprehensive experiments highlight the contribution of modules involved in this retrieval-augmented generation setting, revealing interesting directions for future work.

\end{abstract}
\renewcommand{\thefootnote}
{\fnsymbol{footnote}}
\setcounter{footnote}{2}
\footnotetext{The authors contributed equally to this work.}
\renewcommand{\thefootnote}
{\arabic{footnote}}
\setcounter{footnote}{0}

\section{Introduction}

Text-to-SQL semantic parsing aims at translating natural language questions into SQL, to facilitate querying relational databases by non-experts~\cite{Zelle1996}.
Given their accessibility benefits, Text-to-SQL applications have become popular recently, with many corporations developing Business Intelligence platforms.

The success of Large Language Models (LLMs) in generalising across diverse Natural Language Processing tasks~\cite{Ye2023,OpenAI2024} has fuelled works that looked at how these multi-billion parameter models can be best employed for Text-to-SQL \cite{Liu2023,Pourreza2023}. Recent works in this space have focused on the in-context learning ability of LLMs, demonstrating that significant improvements can be achieved by selecting suitable (question, SQL) example pairs \cite{Nan2023, Gao2023,Guo2024,Sun2024}. 
In spite of its underlying benefits, conventional solutions for example selection are usually limited to retrieving examples based solely on the similarity of questions~\cite{Nan2023, An2023, Guo2024}.
Other approaches resort to a preliminary round of parsing which \textit{approximates} expected SQL queries, and directly use these approximations in few-shot prompting~\cite{Sun2024}, or to subsequently select (question, SQL) pairs by comparing the approximated query to queries within candidate examples \cite{Gao2023}.
The approach proposed by \citeauthor{Gao2023} transforms SQL queries into SQL skeletons \cite{Li2023} and then filters examples by considering overlap token ratio as the similarity between two skeletons. While incorporating SQL skeleton similarity improves over conventional example selection for Text-to-SQL \cite{Gao2023}, it can result in structural information loss as exemplified in Table \ref{tab:sql_skeleton}, where two dissimilar SQL queries are treated as identical. In this paper, we propose a novel approach that selects examples using similarity of normalised SQL Abstract Syntax Trees (ASTs). We argue that considering the similarity of such hierarchical structures can significantly enhance LLMs' performance for Text-to-SQL parsing.

Apart from example selection, we refine database context input to LLMs by dynamically pruning schemata and selecting values.
From the perspective of LLMs, existing studies achieve improvements by including the full database schema in the prompt and additionally \textit{hinting} the importance of particular schema elements or values~\cite{Pourreza2023,Sun2024}. 
In this paper, we show that the performance can be boosted with schemata of reduced size.

Inspired by~\citeauthor{Gao2023} that compute an approximated query for a given input question, we further explore how combinations of a sparse retriever with such an in-parallel semantic parser (we would refer to it as \textit{approximator}) can be used to retrieve relevant database context input to LLMs. For efficiency, we adapt the semantic parser (a decoder-free model with $< 500$M parameters) proposed by \citeauthor{Vougiouklis2023} to process schemata in a parallelised manner.
Using this efficient approximator, our schema pruning strategy  selects a relevant sub-schema in order to simplify the task for LLMs and reduce the relevant computational workload. Furthermore, it enables LLM-based Text-to-SQL solutions to handle longer schemata (usually associated with commercial use-cases) exceeding their context window size.

To realise the above ideas, we proposed a novel approach \astres that features \underline{\textbf{AST}}-based \underline{\textbf{RE}}ranking and \underline{\textbf{S}}chema pruning.
We apply \astres on monolingual (\spider, \spiderdk, \spiderreal and \spidersyn) and cross-lingual (\cspider) benchmarks of different generalisation challenges. We evaluate the applicability of our framework across both closed- and open-source LLMs. Our framework, comprising only a single round of prompting, achieves state-of-the-art performance, outperforming other 
baselines which may comprise complex prompting and multiple iterations, when LLMs of equal capacity are involved. Through comprehensive experiments, we highlight strengths and limitations. Our contributions can be summarised as follows:
\begin{itemize}
    \item We propose a novel approach for selecting (question, SQL) examples using similarity of normalised SQL ASTs.
    \item We take efficient approximation to the extreme, presenting a schema-parallelisable adaptation of the fastest semantic parser to date.
    \item We introduce a framework for dynamically selecting schema elements and database values, offering substantial execution accuracy improvements over prior works while significantly reducing the computational workload of LLMs.
    \item We shed light on the benefits of database value selection and its symbiotic relation to schema pruning for Text-to-SQL LLM prompting.
\end{itemize}

\begin{table}[ht]
\small
\centering
\begin{tabular}{lL{6.cm}}
\toprule
\multirow{2.5}{*}{$\textbf{SQL}_1$} & \texttt{SELECT T2.name, T2.capacity FROM concert AS T1 JOIN stadium AS T2 ON T1.stadium\_id = T2.stadium\_id WHERE T1.year >= 2014}\\ \cmidrule{2-2}
& \textbf{Skeleton:} \texttt{select \_ from \_ where \_} \\
\midrule
\multirow{2.5}{*}{$\textbf{SQL}_2$} & \texttt{SELECT name FROM highschooler WHERE grade = 10}\\ \cmidrule{2-2}
& \textbf{Skeleton:} \texttt{select \_ from \_ where \_} \\

\bottomrule
\end{tabular}
\caption{\label{tab:sql_skeleton}Two SQL queries with identical SQL skeletons.}
\end{table}

\section{Preliminaries}
\label{sec:preliminaries}

Let $\mathbf{q}$ be the sequence of tokens of a natural language question for database $\mathbf{D}$ with tables $\mathbf{t} = t_1, t_2, \ldots, t_{T}$ and columns $\mathbf{c} = c_1^{1}, c_2^{1}, \ldots, c_j^i, \ldots, \allowbreak c_{C_T}^T$, where $c^i_j$ is the $j$-th column of table $t_i$ and $C_i\in \mathbb{N}$ is the total number of columns in table $t_{i}$. Furthermore, let $\mathbf{v_{\mathbf{D}}} = \left\{ v_{c_1^{1}}, v_{c_2^{1}}, \ldots, v_{c_{C_T}^T} \right\}$ be the set of all values associated with the database $\mathbf{D}$ s.t. $v_{c_1^{1}}, \ldots, v_{c_{C_T}^T}$ are the DB value sets associated with respective columns $c_1^{1}, \ldots,  \allowbreak c_{C_T}^T \in \mathbf{c}$.
The goal of the general Text-to-SQL semantic parsing is to predict the SQL query $\mathbf{s}$ given the $(\mathbf{q}, \mathbf{D})$ combination, as follows:
\begin{align}
    \mathbf{s} = \argmax_{\mathbf{s}} p\left(\mathbf{s} \mid \mathbf{q}, \mathbf{D} \right )
\end{align}
For in-context learning, assuming a given LLM of interest, a proxy task for satisfying the original goal involves forming a set of suitable demonstrations, to which we refer as $\mathbb{X^{\star}}$ (see Section~\ref{sec:example_selection}) that would enable the LLM to ad hoc *learn*, how to generate the SQL query that addresses the input $(\mathbf{q}, \mathbf{D})$ combination.
\section{Example Selection using Abstract Syntax Trees}
\label{sec:example_selection}

Our goal is to identify the most suitable set of $\mathbb{X^{\star}} = \left \{ \left ( \mathbf{q}^{\star}_1, \mathbf{s}^{\star}_1 \right ), \ldots, \left ( \mathbf{q}^{\star}_e, \mathbf{s}^{\star}_e \right ) \right \}$ question-SQL pairs from an index of examples, $\mathbb{X}$, s.t. $\mathbb{X^{\star}} \subseteq \mathbb{X}$,  for maximising the probability of an LLM to predict the correct SQL given $(\mathbf{q}, \mathbf{D})$:

\begin{align}
    \mathbb{X^{\star}} = \argmax_{\mathbb{X}} p\left ( \mathbf{s} | \mathbf{q}, \mathbf{D}, \mathbb{X} \right )
\end{align}

From the perspective of ranking, we consider the relevance score between a candidate example $ (\mathbf{q}_j, \mathbf{s}_j) \in \mathbb{X}$ and the input $(\mathbf{q}, \mathbf{D})$. \textit{Vanilla} semantic search is usually based solely on question embeddings, whereas the structure of SQL queries for similar questions is subject to target databases and can thus differ significantly.

To incorporate database context for selecting examples, we propose to re-rank examples retrieved by question embeddings based on normalised SQL ASTs. 
Inspired by~\citeauthor{Gao2023}, our framework utilises a preliminary model to compute an approximated SQL query $\mathbf{s}'$, structurally similar to the ground truth, given $(\mathbf{q}, \mathbf{D})$ s.t. $\mathbf{s}' \sim \mathbf{s}$.

This means that, in an idealised scenario, $\mathbf{s}'$ may be a SQL query that is identical to $\mathbf{s}$, or a query with different lexical aspects (e.g., different variable names or order of \texttt{GROUP BY} operations) but with very high structural similarity to $\mathbf{s}$, such that $\text{score}_{\text{AST}}\left(\mathbf{s}', \mathbf{s}_j \right ) \simeq 1$. Examples are then re-ranked by $\text{score}_{\text{AST}}\left(\mathbf{s}', \mathbf{s}_j \right )$ for each candidate $\mathbf{s}_j$.

AST represents the hierarchical structure of code in a tree form and can be applied to evaluation metrics for code generation~\cite{Tran2019,Ren2020}. The fact that SQL queries sharing identical abstract meanings may not align with the same syntactic structure poses a challenge for measuring similarity through AST differencing.
\paragraph{AST Normalisation} Although it is infeasible to exhaustively transform a SQL to another equivalent form, we can normalise ASTs to reduce undesired mismatch. Firstly, nodes of identifiers are lowercased and unnecessary references are removed (e.g. \verb|<table>.<column>| is substituted with \verb|<column>| if possible). We then delete nodes that create aliases and map each alias to a copy of the subtree to which it references. For cross-domain settings wherein databases at inference time are unseen in the train set, we mask out nodes of values and identifiers after resolving aliases. Otherwise for in-domain settings we further sort nodes associated with JOIN operations(s) to ensure the ordering of tables and keys is consistent.

\paragraph{AST Similarity} Given two normalised ASTs, we adopt the Change Distilling algorithm \cite{Fluri2007} that computes a list of tree edit operations to transform the source AST to the target AST. Types of tree edit operations include: \verb|insert|, \verb|delete|, \verb|alignment|, \verb|move| and \verb|update|. It is essential to note that \verb|move| operation relocates a node to a different parent while moving a node within the same parent is an \verb|alignment|. Therefore, we calculate the similarity between ASTs simply as the ratio of alignments to the total number of operations within the list. Examples of our normalisation and AST similarity are provided in Appendix~\ref{sec:ast_similarity_appendix}.

\section{Database Context Selection}
Apart from relevant question-SQL pairs, prompting for Text-to-SQL parsing requires the context of database schema and values. 

\subsection{Schema Selection}
\label{subsec:schema_retrieval}

We present a hybrid search strategy that selects a sub-schema given a test question to minimise lengthy and potentially irrelevant schema elements 
(i.e. tables and columns) input to LLMs, while maintaining high recall. 

Let $r_{j}^i$ be a semantic representation of column $c_{j}^i$. We aggregate the semantic names\footnote{Semantic name can refer to simply to the name of a particular or to a concatenation of its name and description.} of 
$c_{j}^i$ and the table it belongs to, $t_i$, and its corresponding value set in $\mathbf{D}$, $v_{c_{j}^i}$, as follows:\\
$r_{j}^i = \left \{ t_i \cup c_{j}^i \cup v_{c_{j}^i} \mid i \in \left [1, T \right ] \text{ and }
     j \in  \left [1, C_i \right] 
     \right \}$.

Given question $\mathbf{q}$, we retrieve the most relevant columns using $\text{score}_{\text{BM25}}(\mathbf{q}, r_{j}^i)$ $\forall i \in \left [1, T \right ]$ and $j \in  \left [1, C_i \right]$ \cite{Robertson1994}. A table is retrieved if any of its columns are retrieved.

\subsubsection{Incorporating for Approximated Query}
\label{subsec:approximated_query}
The semantic matching for schema selection requires a comprehension of the relevance between heterogeneous database information (e.g. values and data types) and natural language questions, in addition to interactions across schema elements such as foreign key constraints. 
To this end, a trained parser can inherently be employed as a semantic matching model and elements of a sub-schema are extracted from the approximated query $\mathbf{s}'$.
We argue that a semantic parser which is performing reasonably on the task, can provide us with an $\mathbf{s}'$, whose structure would assimilate the structure of the expected final query, $\mathbf{s}$. Consequently, we opt to dynamically determine the number of columns to be retrieved by $\text{score}_{\text{BM25}}$ as proportional to the number of unique columns in $\mathbf{s}'$, returned by the approximator. A sub-schema is then obtained by merging schema elements selected by $\text{score}_{\text{BM25}}$ with elements from the approximated query.

\subsubsection{Approximating for Longer Schemata}
\label{subsubsec:fastrat_ext}
To further reduce the computational workload, we opt for using a \textit{smaller} model for computing the approximated query, $\mathbf{s}'$. However, smaller models usually have shorter context windows (i.e. $< 2$k tokens), and, as such, they cannot be easily scaled to the requirements of larger schemata. To this end, we propose an approach that enables transformer-based encoders to process longer schemata, in a parallelised manner.

We start with FastRAT \cite{Vougiouklis2023}, which exploits a decoder-free architecture for efficient text-to-SQL parsing. Given a concatenation of the input natural language question $\mathbf{q}$ with the column and table names of a database schema, FastRAT computes the SQL operation in which each element of the input schema would participate in the expected SQL query. We refer to these SQL operations as SQL Semantic Prediction (SSP) labels~\cite{Yu2021}. SQL queries are then deterministically constructed from the predicted SSP labels. We introduce a schema splitting strategy to scale the model up to the requirements of schemata comprising several columns.

We augment the input embedding matrix of the model, with two special schema-completion tokens, \texttt{[full\_schema]} and \texttt{[part\_schema]}, which are used for signalling cases in which a full and a partial schema are provided as input respectively. Our goal is to split a schema consisting of $\sum_{j=1}^{T} C_j$ columns into $r_m$ splits s.t. each split includes a maximum, pre-defined number of columns $r$. Each split consists of the question tokens, a single schema-completion token, the table names of the input $\mathbf{D}$ and up to a maximum $r$ number of columns allocated to this particular split (see Algorithm~\ref{alg:splitting} for further details).

\begin{algorithm}[ht]
\SetKwComment{Comment}{ \# }{ }
\caption{\label{alg:splitting}Algorithm for splitting a schema into smaller splits.} 
\SetKw{Or}{ or }
\KwIn{
$r$ \texttt{: int}\newline
\textcolor{blue}{\texttt{\# concatenation of column}}\newline
\textcolor{blue}{\texttt{\# name tokens}}\newline
$\mathbf{c}^{\text{tok}} \gets [c_1^{\text{tok}_1}, \ldots, c_{C_T}^{\text{tok}_T}]$\newline
\textcolor{blue}{\texttt{\# flatten concatenation of}}\newline
\textcolor{blue}{\texttt{\# table name tokens}}\newline
$\mathbf{t}^{\text{tok}} \gets [t_1^{\text{tok}}, t_2^{\text{tok}}, \ldots, t_{T}^{\text{tok}}]$\newline
\textcolor{blue}{\texttt{\# question tokens}}\newline$\mathbf{q} \gets [q_1, \ldots, q_Q]$\newline

}
\texttt{splits} $\gets \left [ \, \right]$\;
\eIf{$\texttt{len(}\mathbf{c}\texttt{)} > r$}
    {
        \texttt{prefix} $\gets \mathbf{q} + [\textcolor{gray}{\texttt{``[part\_schema]''}}]$\;
    }
    {
        \texttt{prefix} $\gets \mathbf{q} + [\textcolor{gray}{\texttt{``[full\_schema]''}}]$\;
    }

$\texttt{sp} \gets \texttt{prefix}$\Comment*[l]{one sp per split}
\For{$j \gets 1$ to $\sum_{j=1}^{T} C_j$}
    {
        $\texttt{sp} \gets \texttt{sp} + \mathbf{c}^{\text{tok}}\left [ j\right]$\;
        \If{$j\mod r = 0 \Or j=\sum_{j=1}^{T} C_j$}
        {
            $\texttt{sp} \gets \texttt{sp} + \mathbf{t}^{\text{tok}}$\;
            $\texttt{splits.append(sp)}$\;
            $\texttt{sp} \gets \texttt{prefix}$\;
        }
       
    }
\textbf{Return} \texttt{splits}
\end{algorithm}

The returned schema splits along with the SSP labels corresponding to the schema elements of each split are treated as independent instances during training. At test time, an input schema is split according to Algorithm~\ref{alg:splitting}, and the model is input with a batch of the resulting splits. After aggregating the results from all splits, we obtain the SSP label for each column $\in \mathbf{c}$. Inconsistencies across the SSP labels of tables are resolved using majority voting. We refer to this model as $\text{FastRAT}_{\text{ext}}$.

\subsection{Value Selection}
The inference of LLMs for text-to-SQL parsing can be augmented with column values \cite{Sun2024}. We select values for columns in a schema (or a sub-schema) by simply matching keywords in questions and values. This is based on the assumption that LLMs can generalise to unseen values given a set of representative values; thus, the recall and precision of value selection are less critical. We consider value selection providing additional information for LLMs to discern covert differences among columns. An example of our resulting prompt is shown in Appendix~\ref{sec:prompt_formulation}.

\section{Experiments}
We run experiments using two approximators: $\text{FastRAT}_{\text{ext}}$ and Graphix-T5 \cite{Li2023a}. 
Graphix-T5 is is the approximator used by DAIL-SQL~\cite{Gao2023}, and is included to facilitate a fair comparison against the closest work to ours. $\text{FastRAT}_{\text{ext}}$ is trained and tested using $r=64$, unless otherwise stated (see Section~\ref{subsec:schema_splitting}). For all experiments, we use 5 (question, SQL) examples.

We test our approach against both closed- and open-source LLMs: \begin{inparaenum}[(i)]\item \texttt{gpt-3.5-turbo} (\texttt{gpt-3.5-turbo-0613}), \item \texttt{gpt-4} (\texttt{gpt-4-0613}) and \item \texttt{deepseek-coder-33b-instruct}. \end{inparaenum} Results using additional models from the DeepSeek family are provided in Appendix~\ref{subsec:open_source_llm_experiments}.

\subsection{Datasets}

We experiment with several SQL datasets, seeking to explore the effectiveness of our approach on both monolingual and cross-lingual setups. Specifically, we report experiments on \cspider \cite{Min2019} and \spider \cite{Yu2018}.
Since \cspider is a translated version of \spider in Chinese, when it comes to the natural language questions, the characteristics of the two with respect to structure and number of examples are identical.
We focus our evaluation on the development sets\footnote{Appendix~\ref{sec:spider_cspider_exp_appendix} includes results on the test sets.}, which are used as test sets in our experiments. These splits consists of $1,034$ examples of questions on $20$ unique databases that are not met at training time.

We rely on the training splits to maintain an index of (question, SQL) examples, one for each dataset. Using these splits, we train a monolingual and a cross-lingual version of $\text{FastRAT}_{\text{ext}}$.

Furthermore, we use popular \spider variants: \begin{inparaenum}[(i)]\item \spiderdk~\cite{Gan2021a}, \item \spiderreal~\cite{Deng2021} and \item \spidersyn~\cite{Gan2021} \end{inparaenum} to evaluate zero-shot domain generalisation in English (leveraging the \spider (question, SQL) examples index).

Consistently with the relevant leaderboards\footnote{\url{https://taolusi.github.io/CSpider-explorer/} and \url{https://yale-lily.github.io/spider}}, we report results using execution (EX) and exact match (EM) accuracy.\footnote{EX and EM scores are computed using: \url{https://github.com/taoyds/test-suite-sql-eval}.} Since \cspider comes without relevant DB content, we follow previous works, and we focus our evaluation on EM scores \cite{Vougiouklis2023,Cao2023}.

\begin{table}[ht]
\small
\centering
\begin{tabular}{L{5.cm}C{.7cm}C{.7cm}}
\toprule
\textbf{Model} & \textbf{EX} & \textbf{EM}\\\midrule
GraPPa \cite{Yu2021} & $-$ & $73.6$\\
FastRAT \cite{Vougiouklis2023} & $73.2$ & $69.1$\\
$\text{FastRAT}_{\text{ext}}$ & $71.5$ & $64.2$\\ 
Graphix-T5 \cite{Li2023a} & $81.0$ & $77.1$ \\
RESDSQL \cite{Li2023} & $84.1$ & $\mathbf{80.5}$\\\midrule
\multicolumn{3}{l}{\texttt{deepseek-coder-33b-instruct}}\\
\astres (w/ $\text{FastRAT}_{\text{ext}}$) & $81.5$ & $62.1$\\
\astres (w/ Graphix-T5) & $83.4$ & $64.7$\\\midrule
\multicolumn{3}{l}{\texttt{PaLM2}}\\
\textit{Few-shot} SQL-PaLM \cite{Sun2024} & $82.7$ & $-$\\\midrule
\multicolumn{3}{l}{\texttt{text-davinci-003}}\\
Zero-shot \cite{Guo2024} & $73.1$ & $-$\\
RAG w$/$ Rev. Chain \cite{Guo2024} & $\underline{85.0}$ & $-$\\\midrule
\multicolumn{3}{l}{\texttt{gpt-3.5-turbo}}\\
Zero-shot \cite{Liu2023} & $70.1$ & $-$\\
C3 \cite{Dong2023} & $81.8$ & $-$\\
DAIL-SQL \cite{Gao2023} & $79.0$ & $-$\\
\astres (w/ $\text{FastRAT}_{\text{ext}}$) & $82.0$ & $65.7$\\
\astres (w/ Graphix-T5) & $83.0$ & $68.8$\\\midrule
\multicolumn{3}{l}{\texttt{gpt-4}}\\
Zero-shot \cite{Pourreza2023} & $72.9$ & $40.4$\\
DIN-SQL \cite{Pourreza2023} & $82.8$ & $60.1$\\ 
DAIL-SQL \cite{Gao2023} & $83.6$ & $68.7$\\ 
\astres (w/ $\text{FastRAT}_{\text{ext}}$) & $84.3$ & $73.8$\\
\astres (w/ Graphix-T5) & $\textcolor{blue}{\mathbf{86.6}}$ & $\textcolor{blue}{\underline{77.3}}$\\
\bottomrule
\end{tabular}
\caption{\label{tab:spider_res}EX and EM accuracies on the development split of \spider. Fine-tuning-based baselines are listed at the top part of the table. Results of our approach are shown with both $\text{FastRAT}_{\text{ext}}$ and Graphix-T5 as approximators. The \textbf{best} model is in bold, the \underline{second best} is underlined, and the \textcolor{blue}{best prompt-based setup} is in blue.}
\end{table}

\subsection{Baselines}
We compare the performance of our approach against several baselines. We dichotomise the landscape of baselines in fine-tuning- and prompting-based baselines.

\paragraph{Fine-tuning-based} \begin{inparaenum}[(i)] \item \textbf{GraPPa} uses synthetic data constructed via induced synchronous context-free grammar for pre-training an MLM on the SSP-label classification; \item \textbf{DG-MAML} applies meta-learning targeting zero-shot domain generalization; \item \textbf{FastRAT} incorporates a decoder-free framework, by directly predicting SQL queries from SSP labels; \item \textbf{Graphix-T5} inserts a graph-aware layer into T5 \cite{Raffel2020} to introduce structural inductive bias; \item \textbf{RESDSQL} decouples schema linking and SQL skeleton parsing using a framework based on a ranking-enhanced encoder and skeleton-aware decoder; \item \textbf{HG2AST} proposes a framework to integrate dedicated structure knowledge by transforming heterogeneous graphs to abstract syntax trees. \end{inparaenum}

\paragraph{Prompting-based} \begin{inparaenum}[(i)] \item Zero-shot LLM prompting has been explored by \citeauthor{Guo2024,Liu2023,Pourreza2023}; \item \textbf{C3} introduces calibration bias for prompting to alleviate LLMs' biases; \item \textbf{DIN-SQL} uses chain-of-thought prompting with pre-defined prompting templates tailored for the assessed question hardness; \item \textbf{DAIL-SQL} uses  query approximation and SQL skeleton-based similarities for example selection; \item \textbf{SQL-PaLM} proposes a framework for \textit{soft} column selection and execution-based refinement; \item \textbf{RAG w$\mathbf{/}$ Rev. Chain} augments the input prompt with question skeleton-based example retrieval and an execution-based revision chain.\end{inparaenum}

\begin{table*}[ht]
    \begin{minipage}{.99\textwidth}
    \small
    \centering
    \begin{tabular}{l c c c c  c c c}
    \toprule
    \multirow{2.5}{*}{\textbf{Model}} & \multicolumn{2}{c}{\textbf{\spiderdk}} & \multicolumn{2}{c}{\textbf{\spiderreal}} & \multicolumn{2}{c}{\textbf{\spidersyn}} & \textbf{\cspider} \\\cmidrule{2-8} 

         & \textbf{EX} & \textbf{EM} & \textbf{EX} & \textbf{EM} & \textbf{EX} & \textbf{EM} & \textbf{EM} \\ \midrule
    RAT-SQL + BERT \cite{Wang2020} & $-$  & $40.9$ & $62.1$ & $58.1$ & $-$ & $48.2$ & $-$ \\
    DG-MAML \cite{Wang2021} & $-$ & $-$ & $-$ & $-$ & $-$ & $-$ & $51.0$\\
    FastRAT \cite{Vougiouklis2023} & $-$ & $-$ & $-$ & $-$ & $-$ & $-$ & $\underline{61.3}$\\
    $\text{FastRAT}_\text{ext}$ & $-$ & $44.1$  & $-$ & $47.8$ & $-$ & 48.5 & $53.2$ \\
    HG2AST \cite{Cao2023} & $-$ & $-$ & $-$ & $-$ & $-$ & $-$ & $61.0$\footnote{Without using question translation; $64.0$ EM when question translation is used.}\\
    RESDSQL \cite{Li2023} & $66.0$ & $\underline{55.3}$ & $\mathbf{81.9}$ & $\mathbf{77.4}$ & $\mathbf{76.9}$ & $\mathbf{69.1}$ & $-$ \\ \midrule
    \multicolumn{7}{l}{\texttt{deepseek-coder-33b-instruct}}\\
    \astres (w/ $\text{FastRAT}_{\text{ext}}$) & $\underline{70.5}$ & $46.4$ & $77.4$ & $59.3$ & $68.7$ & $49.5$ & $55.9$\\\midrule
    \multicolumn{7}{l}{\texttt{gpt-3.5-turbo}}\\
    Zero-shot \cite{Liu2023} &  $62.6$ & $-$  & $63.4$ & $-$ & $58.6$ & $-$ & $32.6$ \\
    DAIL-SQL \cite{Gao2023} & $-$ & $-$ & $67.9$ & $-$ & $-$ & $-$ & $-$\\
     \astres (w/ $\text{FastRAT}_\text{ext}$) & $68.8$ & $49.3$ & $78.0$ & $60.8$ & $66.9$ & $51.2$ & $54.0$ \\
    \midrule
    \multicolumn{3}{l}{\texttt{PaLM2}}\\
    \textit{Few-shot} SQL-PaLM \cite{Sun2024} & $66.5$ & $-$ & $77.6$ & $-$ & $\textcolor{blue}{\underline{74.6}}$ & $-$ & $-$ \\\midrule

    \multicolumn{7}{l}{\texttt{gpt-4}}\\
    DAIL-SQL \cite{Gao2023} & $-$ & $-$ & $76.0$ & $-$ & $-$ & $-$ & $-$\\
    \astres (w/ $\text{FastRAT}_\text{ext}$) & $\textcolor{blue}{\mathbf{72.3}}$ & $\textcolor{blue}{\mathbf{59.1}}$ & $\textcolor{blue}{\underline{80.9}}$ & $\textcolor{blue}{\underline{66.1}}$ & $74.4$ & $\textcolor{blue}{\underline{61.3}}$ & $\textcolor{blue}{\mathbf{64.4}}$ \\
    \bottomrule
    
     \end{tabular}
    \caption{Results on \spiderdk, \spiderreal, \spidersyn and \cspider. Fine-tuning-based baselines are listed at the top. The \textbf{best} model is in bold, \underline{second best} is underlined, and the \textcolor{blue}{best prompt-based setup} is in blue.
    }
    \label{tab:spider_variants_cspider}
    \end{minipage}
\end{table*}

\subsection{Text-to-SQL Evaluation}
Table~\ref{tab:spider_res} and \ref{tab:spider_variants_cspider} summarise the results of our approach with \texttt{deepseek-coder-33b-instruct}, \texttt{gpt-3.5-turbo} and \texttt{gpt-4} against the baselines. Our approach, comprising a single-prompting round, surpasses other LLM-based solutions, that incorporate several prompting iterations, for LLMs of the same capacity. We note consistent improvements over DAIL-SQL, the closest work to ours, even when $\text{FastRAT}_{\text{ext}}$ is used as approximator (i.e. a model consisting of $< 500$M vs the $\geq 3$B parameters that DAIL-SQL's approximator is using). For the same approximator, our framework is able to meet, performance standards of DAIL-SQL (equipped with \texttt{gpt-4} and an additional self-consistency prompting step) using an open-source model as backbone LLM, by achieving shorter prompts in a single prompting step. 

\spider results are consistent with the results across the various Spider variants and \cspider\footnote{In \cspider, questions are fully translated in Chinese while the DB content remains in English. Due to this limitation, DB schema and content selection are disabled.} (Table~\ref{tab:spider_variants_cspider}).
Our approach levering $\text{FastRAT}_{\text{ext}}$ and AST-based re-ranking for example selection out-performs other prompting-based solution, and is in-line with the scores of state-of-the-art fine-tuning-based baselines. While \texttt{gpt-4} is the most capable model within our framework (with this being more noticeable in the case of \spidersyn), we observe surprising findings with DeepSeek with which in many cases our approach can surpass much more computationally expensive alternatives based on larger closed-source LLMs. Our findings remain consistent with \cite{Liu2023} since the EM scores of prompting-based methods fall behind those of their fine-tuning based counterparts.

\subsubsection{Schema Selection Evaluation}

\begin{table*}[ht]
\small
\centering
\begin{tabular}{l l c c c c}
\toprule
\textbf{Approximator} & \textbf{Schema Selection Setup} & \textbf{Recall} & \textbf{Schema Shorten.} & \textbf{EX} & \textbf{EM} \\\midrule

\textit{Oracle} & Gold Query & $100.0$ & $71.3$ & $86.3$ & $73.9$\\\midrule
$\text{FastRAT}_{\text{ext}}$ & N$/$A & $100.0$ & $0.0$ & $79.3$ & $63.6$\\
$\text{FastRAT}_{\text{ext}}$ & BM25 (top-$10$) & $92.0$ & $36.5$ & $78.9$ & $64.1$ \\
$\text{FastRAT}_{\text{ext}}$ & BM25 (top-$20$) & $98.3$ & $14.1$ & $80.7$ & $64.9$ \\
$\text{FastRAT}_{\text{ext}}$ & Approx. Query & $86.8$ & $71.3$ & $78.4$ & $63.8$ \\
$\text{FastRAT}_{\text{ext}}$ & Approx. Query + BM25 (top-$7$) & $93.3$ & $50.4$ & $81.1$ & $65.4$\\
$\text{FastRAT}_{\text{ext}}$ & Approx. Query + BM25 (top-$10$) & $97.0$ & $37.3$ & $81.2$ & $65.6$ \\
$\text{FastRAT}_{\text{ext}}$ & Approx. Query + BM25 (dynamic top-$k$) & $97.2$ & $49.0$ & $\mathbf{82.0}$ & $\mathbf{65.7}$\\

\midrule
Graphix-T5 & N$/$A & $100.0$ & $0.0$ & $79.8$ & $65.6$\\
Graphix-T5  & Approx. Query & $92.3$ & $71.8$ & $81.8$ & $68.8$ \\
Graphix-T5  & Approx. Query + BM25 (dynamic top-$k$) & $97.9$ & $49.4$ & $\mathbf{83.0}$ & $\mathbf{68.8}$\\
\bottomrule
\end{tabular}
\caption{\label{tab:recall_compres}Recall, Schema Shortening, EX and EM scores (using \texttt{gpt-3.5-turbo}) across different schema selection setups, on the development split of \spider. Value selection is enabled for the selected columns across all setups. For the oracle setup, we report performance upper-bounds using \textit{only} the schema elements from the gold query.}
\end{table*}
We evaluate our proposed schema selection strategy in a two-fold manner, given that value selection is applied for selected columns. Firstly, we use recall and schema shortening (rate) to compute averaged metrics across all samples showcasing the extent to which \begin{inparaenum}[(i)] \item the most relevant schema elements are successfully retrieved, and \item the size of the resulting schema, after selection, with respect to its original size\end{inparaenum}. Secondly, we explore how the performance of the end-system changes across different schema pruning settings by reporting EX and EM scores. Recall is the percentage of samples for which all ground-truth schema elements are selected. Schema shortening is the number of schema elements that are excluded divided by the total number of schema elements. Results are summarised in Table~\ref{tab:recall_compres}.

We consider results of using only gold queries to be the upper bound. In this setup, the highest execution accuracy is achieved while filtering out $>70\%$ of the original schema on average. We note that our approach of coupling the schema elements returned in the approximated query with the ones returned by BM25 navigates a healthy trade-off between maximising recall and reducing processing of unnecessary schema elements. We also notice that our strategy of dynamically determining the number of retained schema elements per input (see Section~\ref{subsec:approximated_query}) results in improvements compared to static top-$k$ determination. For roughly the same extent of schema shortening (i.e. by comparing scores with dynamic top-$k$ against top-$7$), the results with the former are higher across all metrics. As discussed in Section \ref{subsec:approximated_query}, the semantic matching for schema selection is non-trivial and additional experiments illustrating the efficacy of using an approximator in comparison to a dense retriever are provided in Appendix \ref{sec:extra_schema_selection_appendix}.

\subsubsection{Ablation Study}
\label{subsubsec:example_sel_evaluation}
Table~\ref{tab:approx_ast_svs} shows a comprehensive ablation study for the efficacy of our database context selection, and example selection methods including DAIL \cite{Gao2023} and AST.
We consistently notice improvement when selecting examples using AST, for the same approximator. Interestingly, the performance gap is increasing the better the approximator becomes, leading to an improvement $> 2.4\%$ in the case of an oracle approximator. This finding is in agreement with our hypothesis that AST re-ranking can preserve structural information for more precise example selection when $\mathbf{s}' \sim \mathbf{s}$. The inclusion of combined schema and value selection (SVS) leads to further improvements when coupled with example selection based on AST or DAIL.

\begin{table}[ht]
\small
\centering
\begin{tabular}{clcc}
\toprule
\textbf{Approximator} & \textbf{Selection}  & \textbf{EX} & \textbf{EM} \\ \midrule
 N$/$A & Question Similarity & 74.7 & 52.3 \\ 
\midrule

\multirow{6}{*}{$\text{FastRAT}_{\text{ext}}$} & DAIL  & 78.6 & 61.4 \\
& DAIL + SVS & 81.3 & 62.3 \\
& AST & 79.3 & 63.6 \\
& AST + VS & 80.4 & 63.8 \\
& AST + SS & 78.9 & 63.8 \\
& AST + SVS & 82.0 & 65.7 \\
\midrule

\multirow{6}{*}{Graphix-T5} & DAIL & 77.8 & 61.9 \\
& DAIL + SVS & 81.0 & 63.7 \\
& AST & 79.8 & 65.6 \\
& AST + VS & 81.4 & 66.4 \\
& AST + SS & 80.2 & 66.2 \\
& AST + SVS & 83.0 & 68.8 \\
\midrule

\multirow{6}{*}{\textit{Oracle}} & DAIL  & 79.1 & 63.2 \\
& DAIL + SVS & 82.5 & 66.0 \\
& AST & 81.0 & 67.6 \\
& AST + VS & 82.6 & 68.1 \\
& AST + SS & 82.5 & 69.6 \\
& AST + SVS & 84.6 & 71.3 \\
\bottomrule

\end{tabular}
\caption{\label{tab:approx_ast_svs}EX and EM scores on the development set of \spider, with \texttt{gpt-3.5-turbo}, across different approximators, and selection setups: example selection (with DAIL or AST), schema selection (SS), value selection (VS), and schema \& value selection (SVS). We report results from oracle approximator using gold queries.}

\end{table}

\subsection{Schema Splitting}
\label{subsec:schema_splitting}
\begin{figure}[ht]
  \centering
  \includegraphics[width=0.95\linewidth]{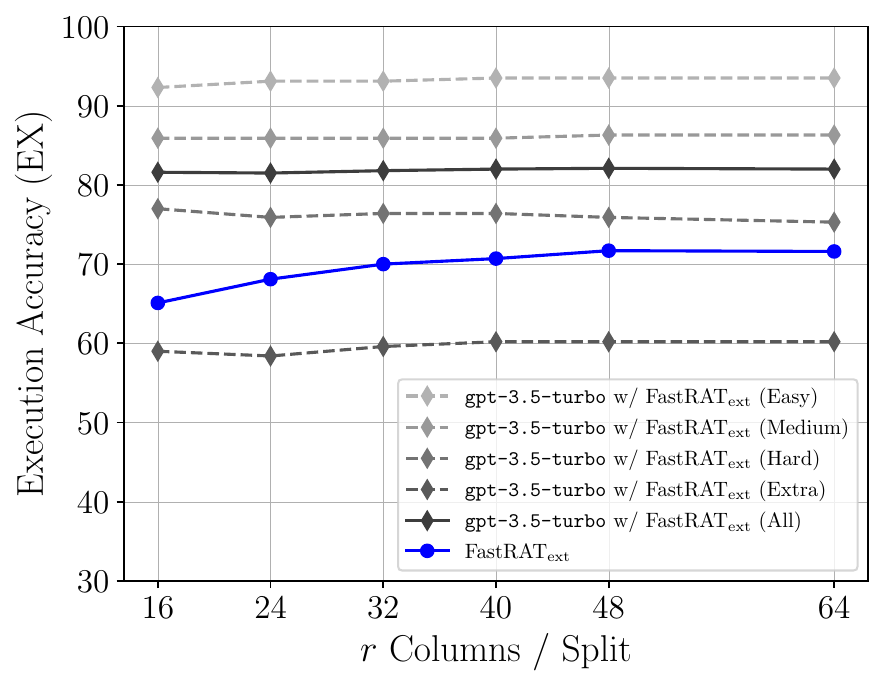}
  \caption{\label{fig:exec_accuracy_across_splits}Execution accuracy scores on on the development set of \spider across different maximum numbers of columns per schema split, $r$. The results of our approach, using \texttt{gpt-3.5-turbo}, are presented across different \spider-query difficulty levels.}
  
\end{figure}

We evaluate the effect of splitting a schema into $r_m$ splits, using $\text{FastRAT}_\text{ext}$ for schema selection. Figure~\ref{fig:exec_accuracy_across_splits} shows EX scores across different maximum number of columns per schema split ($r$), on the development set of \spider. We see that the EX scores of our approach remain consistent across different $r$. The performance in the case where particular schemata from the development set are split into $r_m=3$ or $r_m=4$ splits (i.e. for $r=24$ or $r=16$ respectively) is identical to the scores where schemata are split using the default $r$ with which $\text{FastRAT}_\text{ext}$ has been trained.

\section{Discussion}
\label{ref:discussion}
\paragraph{Are there any theoretical performance upper limits for example selection using AST?} For each data instance in the development set of Spider, we compute the average AST similarity between the approximated query and each SQL query that is included (after example selection) in the corresponding prompt. In Table~\ref{tab:ast_eg_sufficiency}, we measure EX scores on the development set of \spider, across different AST-similarity intervals. We see an obvious correlation between execution accuracy and AST scores--execution accuracy is higher for higher AST scores. Besides highlighting an empirical, execution accuracy upper-bound, in the case of test questions whose SQL structure is well-covered (i.e. with high AST score) in the examples space, our approach can hint data instances that might be challenging for the current configuration, without even requiring to prompt the target LLM or executing the resulting SQL against a database instance. 
Challenging data instances can be taken into consideration with respect to the existence of insufficient examples in $\mathbb{X}$ to support the expected SQL structure or a potentially harmful approximator.

\begin{table}[ht]
    \centering
    \small
    \begin{tabular}{l c c c}
    \toprule
    \multirow{2.5}{*}{\textbf{AST Interval}} & \multicolumn{3}{c}{\textbf{Approximator}} \\ \cmidrule{2-4}

         & \textit{Oracle} & Graphix & $\text{FastRAT}_\text{ext}$  \\ \midrule
    $[0.0, 1.0]$ & $84.6$  & $83.0$ & $82.0$ \\ \midrule
    $[0.95, 1.0]$ & $90.8$  & $88.4$ & $88.7$ \\
    $[0.9, 0.95)$ & $80.7$ & $74.1$ & $76.8$  \\
    $[0.85, 0.9)$ & $73.0$ & $68.3$ & $59.4$ \\
    $[0.8, 0.85)$ &  $62.4$ & $66.8$ & $63.5$ \\
    $[0.0, 0.8)$ & $50.0$ & $54.1$ & $58.7$ \\
    \bottomrule
    
     \end{tabular}
    \caption{EX scores on the \spider development set using \texttt{gpt-3.5-turbo}, across different average AST-similarity intervals.}
    \label{tab:ast_eg_sufficiency}
\end{table}

\paragraph{Is the choice of approximator critical?} Although our AST re-ranking and schema selection benefit from more accurate SQL predicted by a stronger approximator, the choice of approximator depends on the desired trade-off between effectiveness and efficiency in practice. $\text{FastRAT}_{\text{ext}}$ is over $600$ times faster than Graphix-T5 \cite{Li2023a} on an A100 $80$G GPU, while the resulting difference, within our framework, in EX on the \spider development set is $\leq 1\%$ (Table \ref{tab:approx_ast_svs}). 

\paragraph{Does schema selection improve the performance?}
In Table~\ref{tab:approx_ast_svs}, we noted that performing schema selection (i.e. SS) without DB value selection does not necessarily lead to performance improvements. This is in partial agreement with \citeauthor{Sun2024} that hard column selection can be harmful for the end-to-end performance, and can be attributed to the drop of recall, when less capable approximators are involved. Nonetheless, as we note in Section~\ref{subsubsec:example_sel_evaluation}, the combination of schema and value selection (SVS) can consistently improve EX and EM, while significantly reducing the LLM token-processing cost due to shortened schema.

\section{Related Work}
\label{sec:background}

Significant number of recent works have looked at how LLMs can be employed in Text-to-SQL scenarios \cite{Rajkumar2022,Chang2023, Liu2023,Pourreza2023,Gao2023,Guo2024}.
More recent works have looked at how incorporating examples in the prompt could benefit the performance of LLMs in the end task~\cite{Pourreza2023,Gao2023,Guo2024,Sun2024}. 
In spite of its underlying benefits, conventional solutions for example selection have focused on retrieving pairs using question similarity~\cite{Nan2023}. Other approaches have sought to \textit{approximate} expected SQL queries, and either directly use these approximations in the prompt, in a few-shot setting~\cite{Sun2024} or to filter candidate (question, SQL) pairs by taking into consideration the similarity of their corresponding SQL query skeletons \cite{Li2023} against the skeleton of the approximated SQL~\cite{Gao2023}. We argue that such example selection strategies can result in information loss, and we propose an approach for re-ranking examples using similarity of normalised SQL ASTs.

Retrieving (question, SQL) pairs using tree-edit distances have been recently explored in the context of synthesising parallel datasets \cite{Awasthi2022} and few-shot settings for adapting models to a target schema at inference time~\cite{Varma2023}. In our work, we introduce a more refined normalisation paradigm for AST that goes beyond conventional considerations for making the resulting trees invariant to the mentions of database schema elements and values \cite{Awasthi2022}. Furthermore, in contrast to CTreeOT by \citeauthor{Varma2023}, we focus on the cross-domain setting, without relying on the availability of a handful of in-domain examples for the test schemata.

The benefits of schema selection for Text-to-SQL have been highlighted across the relevant bibliography~\cite{Wang2020,Li2023a,Pourreza2023}. From the LLMs perspective, pruning schema elements from the prompts has been usually leading to performance degradation \cite{Sun2024}.
Inspired by~\citeauthor{Gao2023}, we compute a preliminary query for a given $(\mathbf{q}, \mathbf{D})$ by we adapting FastRAT \cite{Vougiouklis2023}, to the requirements of processing longer schemata, in a parallelised manner. We couple the resulting \textit{approximator} with a sparse retriever, and we propose a dynamic strategy for reducing the computational cost of the task while achieving performance improvements.

\section{Conclusion}
\label{sec:conclusion}

In this paper, we propose \astres, a flexible yet powerful framework that augments LLMs for Text-to-SQL semantic parsing by selecting suitable examples and database information. We present a novel AST-based metric to rank examples by similarity of SQL queries. Our hybrid search strategy for schema selection reuses a preliminary query to reduce irrelevant schema elements while maintaining high recall. Extensive experiments demonstrate that our AST-based ranking outperforms previous approaches of example selection and that a symbiotic combination of schema and value selection can further enhance the end-to-end performance of both closed- and open-source LLM solutions.

\section*{Limitations}
There are limitations with regards to both of our example selection and schema selection. Our AST-based ranking can be biased when an approximated SQL deviates significantly from structurally correct answers. To address the failure of approximators, a future direction is to sensibly diversify selected examples such that LLMs can generalise compositionally. As for schema selection, our semantic search relies on an approximator which is essentially a parser with high precision in schema linking but lack mechanisms to control recall as a standalone model. Therefore, it is worth extending cross-encoder architecture such as $\text{FastRAT}$ to support ranking schema elements while being a SQL approximator in the meantime.

We demonstrate that schema splitting strategies within our framework can be applied across various numbers of splits without noticeable performance degradation. Nonetheless, given the lack of available datasets that incorporate longer commercial schemata, we focus our experiments on the cross-database setting provided by \cspider and \spider variants.

\section*{Ethics Statement}

We do not make use of any private, proprietary, or sensitive data. $\text{FastRAT}_{\text{ext}}$ is trained on publicly available Text-to-SQL datasets, using publicly available encoder-models as base. Our framework for retrieval-augmented generation builds on-top of large, pre-trained language models, which may have been trained using proprietary data (e.g. in the case of the OpenAI models). Given the nature of pre-training schemes, it is possible that our system could carry forward biases present in the datasets and/or the involved LLMs.

\bibliography{main}
\appendix

\section{SQL Similarity using Normalised Abstract Syntax Trees}
\label{sec:ast_similarity_appendix}

\begin{figure}[ht]
  \centering
  \includegraphics[width=0.95\linewidth]{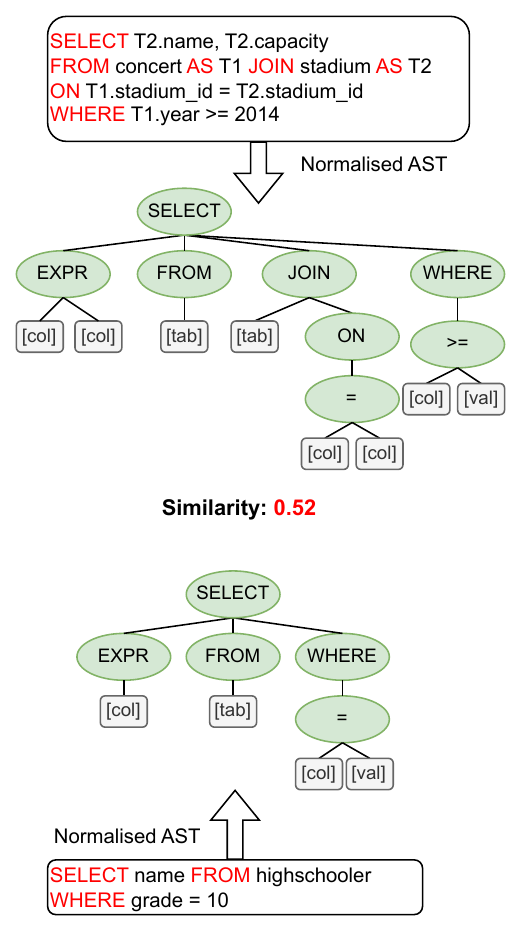}
  \caption{\label{fig:ast_similarity}Example of how the similarity between two different SQL queries is computed using normalised ASTs.}
  
\end{figure}

Table \ref{tab:ast_normalisation} shows the corresponding SQL queries after each step of our AST normalisation as explained in Section \ref{sec:example_selection}. An example of the similarity between normalised ASTs is provided in Figure~\ref{fig:ast_similarity}, where tables, columns and values are masked out for cross-domain settings.

\begin{table}[ht]
\small
    \centering
    \begin{tabular}{L{1.45cm}L{5.45cm}}
    \toprule
    \textbf{SQL}   & \texttt{\red{SELECT} T1.Category, \red{COUNT}(*) \red{AS} Num \red{FROM} Products \red{AS} T1 \red{JOIN} Orders \red{AS} T2 \red{ON} T1.id = T2.pid \red{GROUP BY} T1.Category \red{ORDER BY} Num \red{ASC}}  \\ \midrule
    \textbf{1. Unify Identifiers}  & \texttt{\red{SELECT} t1.category, \red{COUNT}(*) \red{AS} num \red{FROM} products \red{AS} t1 \red{JOIN} orders \red{AS} t2 \red{ON} t1.id = t2.pid \red{GROUP BY} t1.category \red{ORDER BY} num \red{ASC}} \\ \midrule
    \textbf{2. Resolve Aliases}  & \texttt{\red{SELECT} products.category, \red{COUNT}(*) \red{FROM} products \red{JOIN} orders \red{ON} products.id = orders.pid \red{GROUP BY} products.category \red{ORDER BY} \red{COUNT}(*) \red{ASC}} \\ \midrule
    \textbf{3. Reorder JOIN} & \texttt{\red{SELECT} products.category, \red{COUNT}(*) \red{FROM} orders \red{JOIN} products \red{ON} orders.id = products.pid \red{GROUP BY} products.category \red{ORDER BY COUNT}(*) \red{ASC}} \\ \midrule
    \textbf{4. Mask Identifiers \& Values} (\textcolor{blue}{cross-domain only})  & \texttt{\red{SELECT} \_, \red{COUNT}(*) \red{FROM} \_ \red{JOIN} \_ \red{ON} \_ = \_ \red{GROUP BY} \_ \red{ORDER BY COUNT}(*) \red{ASC}} \\
    
    \bottomrule
    \end{tabular}
    \caption{An example of the effects to corresponding SQL after each step of our AST normalisation.}
    \label{tab:ast_normalisation}
\end{table}

\begin{table}[ht]
    \centering
    \scriptsize
    \begin{tabular}{L{1cm}|p{5.8cm}}
    \toprule
     & \# Given SQLite database schema student\_transcripts:\\
    \parbox{0pt}{\colorbox{pale-orange}{Selected}\\\colorbox{pale-orange}{Schema} w$/$ \colorbox{pink}{Selected}\\ \colorbox{pink}{Values}} & \cellcolor{pale-orange}\parbox{0.95\linewidth}{%
\texttt{CREATE TABLE Departments(\\
\hspace*{2em}department\_id number,\\
\hspace*{2em}department\_name text COMMENT 'department name \colorbox{pink}{(e.g. engineer, statistics, medical)}', \\
\hspace*{2em}...);\\
CREATE TABLE Degree\_Programs(\\
\hspace*{2em}degree\_program\_id number,\\
\hspace*{2em}degree\_summary\_name text COMMENT 'degree summary name \colorbox{pink}{(e.g. PHD, Master, Bachelor)}',
\hspace*{2em}...\\
\hspace*{2em}PRIMARY KEY (degree\_program\_id), \\
\hspace*{2em}FOREIGN KEY (department\_id) REFERENCES Departments(department\_id));}
}\\
& \# Your task is to translate Question into SQL.\\
& \# Some examples are provided based on similar problems:\\
\parbox{0pt}{\colorbox{pale-green}{Selected}\\
\colorbox{pale-green}{Examples}} &
\cellcolor{pale-green}\parbox{0.95\linewidth}{
Question: How many courses does the department of Computer Information Systems offer? \raggedright\\
SQL: \texttt{SELECT count(*) FROM department AS T1 JOIN course AS T2 ON T1.dept\_code = T2.dept\_code WHERE dept\_name = ``Computer Info.Systems''}\\
Question: ...\\
SQL: ...}\\
& \# Complete the following SQL for schema student\_transcripts:\\
\parbox{0pt}{\colorbox{pale-blue}{Test}\\\colorbox{pale-blue}{Question}} & \cellcolor{pale-blue}\parbox{0.95\linewidth}{
Question: How many degrees does the engineering department have?\\
SQL: 
}\\
    \bottomrule
    \end{tabular}
    \caption{\label{tab:prompt_formulation}An example of the resulting prompt, after example and schema and DB content selection.}
    
\end{table}

\section{Prompt Formulation}
\label{sec:prompt_formulation}
Table \ref{tab:prompt_formulation} shows an example of our prompt, after example and DB context selection (i.e. schema and value selection). This prompt is provided as input to LLMs. Following the latest OpenAI example\footnote{\url{https://platform.openai.com/examples/default-sql-translate}} for Text-to-SQL parsing, we represent a schema with \texttt{CREATE TABLE} statements in SQL. Semantic names or descriptions of tables and columns are included as \texttt{COMMENT} along with the corresponding columns or tables. Note that we 
filter out comments that can be obtained by simply lowercasing original names and/or removing underscores. To maintain a compact representation of database information, we append selected values of columns into their \texttt{COMMENT} rather than introducing additional lines as in the work by \citet{Chen2024}. Example (question, SQL) pairs are provided in a similar manner to DAIL-SQL \cite{Gao2023}, followed by an instruction to prompt LLMs to generate SQL for the test question.

\section{Implementation Details}
\label{sec:implementation_details}
We use this section to provide further details about the implementation of our approach.
\subsection{Example Selection}
\label{subsec:implementation_details_exp_select}
Following \cite{Gao2023}, we employ the pre-trained \texttt{all-mpnet-base-v2} model~\cite{Song2020} from Sentence Transformer~\cite{Reimers2019} to compute dense question embeddings for English datasets including \spider, \spiderdk, \spiderreal, and \spidersyn. For \cspider, \texttt{paraphrase-multilingual-MiniLM-L12-v2} is used instead. SQL queries are parsed into AST by using SQLGlot\footnote{\url{https://github.com/tobymao/sqlglot}} and are then normalised as explained in Section \ref{sec:example_selection}. SQLGlot provides an implementation of the Change Distilling algorithm for AST differencing. We refer readers to SQLGlot's documentation\footnote{\url{https://github.com/tobymao/sqlglot/blob/main/posts/sql_diff.md}} for more details. For selecting example question-SQL pairs, we first retrieve top 500 examples by question similarity and rerank them in terms of the similarity of normalised SQL ASTs. For relevant experiments in Table \ref{tab:approx_ast_svs}, we reproduced the implementation of DAIL selection from the original paper \cite{Gao2023}. The number of few-shot examples is set to 5 across all experiments.

\subsection{Schema \& Value Selection}
\label{subsec:schema_val_select_appendix}
Each database schema is treated as an independent collection of columns that are analogous to documents to be retrieved by using BM25.
As mentioned in Section \ref{subsec:schema_retrieval}, we represent a column by concatenating semantic names of both the column and its table, and the column values in the database.
Semantic names and values are tokenized using spaCy\footnote{\url{https://github.com/explosion/spaCy}} and preprocessed by lowercasing and stemming\footnote{\url{https://www.nltk.org/api/nltk.stem.porter.html}}. At inference time, the same processing is applied to questions. We adopt the implementation of Okapi BM25 \cite{Robertson1994} from Rank-BM25\footnote{\url{https://github.com/dorianbrown/rank_bm25}}.
The number of columns to retrieve is dynamically set to $\lfloor 1.5 \times \gamma \rfloor$ where $\gamma$ is the number of unique columns in an approximated query.
We limit the resulting number to a range between 6 and 20. 
By retrieving at column level, a table is selected if any of its columns are selected. We merge retrieved schema elements with schema elements from the approximated query to construct a sub-schema.
To further increase the recall, we add additional primary keys and foreign keys that are not selected but valid based on selected tables, except for experiments where only approximated queries are used (see Table \ref{tab:recall_compres}). In such cases, however, if the SQL query involves only tables (e.g. \texttt{SELECT * FROM books}), primary keys of selected tables are still included to ensure that corresponding \texttt{CREATE TABLE} statements (see Table \ref{tab:prompt_formulation}) are meaningful and consistent.

For selecting values, similarly, we match the input question and the set of values for each (selected) column that has a non-numeric type. The top 3 results are added to the prompt as exemplified in Table \ref{tab:prompt_formulation}. The same setting of schema and value selection is used for all datasets we experimented with except \cspider. Due to the cross-lingual nature of \cspider, schema selection and value selection are simply disabled.

\paragraph{Training \texorpdfstring{$\text{FastRAT}_\text{ext}$}{Lg}}We follow the original hyper-parameters provided by~\cite{Vougiouklis2023} for training $\text{FastRAT}_\text{ext}$. The monolingual version of FastRAT is based on $\text{BERT}_{\text{LARGE}}$ while its cross-lingual variant on XLM-RoBERTa-large.

\subsection{OpenAI Models}
We use \texttt{gpt-4} (\texttt{gpt-4-0613}) and \texttt{gpt-3.5-turbo} (\texttt{gpt-3.5-turbo-0613}) for our experiments. For decoding, sampling is disabled and the maximum number of tokens to generate is set to 256. A single experiment, on the \spider development set using our approach with $\text{FastRAT}_\text{ext}$ as the approximator and dynamic database context selection costs around $\$0.8$ and $\$16.5$ in the case of \texttt{gpt-3.5-turbo-0613} and \texttt{gpt-4-0613} respectively.

\subsection{Experiments with Open-Source LLMs}
\label{subsec:open_source_llm_experiments}
We further conduct experiments with open-source models from the DeepSeek family\footnote{We use the implementations provided by \url{https://huggingface.co/deepseek-ai}.}, that specialise in code generation. Prompting and decoding setups remain consistent across all LLMs. Table~\ref{tab:deepseek_results} summarises the results. We see that our approach can generalise even in the case of open-source LLM alternatives. Interestingly, our scores using \texttt{deepseek-coder-33b-instruct} are comparable to the scores when using \texttt{gpt-3.5-turbo-0613} across all approximators. Inference experiments are conducted on a machine using $8 \times$NVIDIA-V100 32G GPUs.

\begin{table*}[t]
\small
\centering
\begin{tabular}{lcccccccccc}
\toprule
\multirow{2.5}{*}{\textbf{Model}} & \multicolumn{2}{c}{\textbf{\spider}} & \multicolumn{2}{c}{\textbf{\spiderdk}} & \multicolumn{2}{c}{\textbf{\spiderreal}} &
\multicolumn{2}{c}{\textbf{\spidersyn}} & \textbf{\cspider} \\ \cmidrule{2-10}
& \textbf{EX} & \textbf{EM} & \textbf{EX} & \textbf{EM} & \textbf{EX} & \textbf{EM} & \textbf{EX} & \textbf{EM} & \textbf{EM} \\\midrule
\multicolumn{3}{l}{\texttt{deepseek-coder-6.7b-instruct}}\\
\astres (w/ $\text{FastRAT}_{\text{ext}}$) & $78.6$ & $64.8$ & $66.7$ & $46.4$ & $73.2$ & $55.7$ & $66.0$ & $49.5$ & $53.0$ \\
\astres (w/ Graphix-T5) & $79.5$ & $64.8$ & $-$ & $-$ & $-$ & $-$ & $-$ & $-$ & $-$ \\\midrule
\multicolumn{3}{l}{\texttt{deepseek-coder-33b-instruct}}\\
\astres (w/ $\text{FastRAT}_{\text{ext}}$) & $81.5$ & $62.1$ & $70.5$ & $46.4$ & $77.4$ & $59.3$ & $68.7$ & $49.5$ & $55.9$\\
\astres (w/ Graphix-T5) & $83.4$ & $64.7$ & $-$ & $-$ & $-$ & $-$ & $-$ & $-$ & $-$
\\
\bottomrule
\end{tabular}
\caption{\label{tab:deepseek_results}Execution (EX) and exact match (EM) accuracy scores of our approach using DeepSeek family models, on the development splits of \spider and \cspider, and the \spiderdk, \spiderreal and \spidersyn test splits. \cspider results are using only $\text{FastRAT}_{\text{ext}}$ as approximator.}
\end{table*}

\section{\spider and \cspider Experiments}
\label{sec:spider_cspider_exp_appendix}
We report experiments on \cspider \cite{Min2019} and \spider \cite{Yu2018}, which contain database schema information and examples in Chinese and English respectively. 
Since \cspider is a translated version of the \spider dataset, the characteristics of the two with respect to structure and number of examples are identical. Both datasets contain $8,659$ examples of questions and SQL queries along with their relevant SQL schemata (i.e. $146$ unique databases). The development and test\footnote{Since the $1$st of March $2024$, the test sets of both Spider and CSpider have become publicly available.} sets consist of $1,034$, on $20$ unique databases and $2,147$, on $40$ unique databases, respectively, and none of the relevant databases are seen in the training set. Due to the scarcity of works reporting test scores on these benchmarks, we chose not to include our results in the main body of our manuscript. Table~\ref{tab:spider_test_scores} shows the performance of our framework with respect to execution and exact match accuracy scores on the test splits of \spider and \cspider.

\begin{table*}[t]
\small
\centering
\begin{tabular}{lccccccccccc}
\toprule
\multirow{2.5}{*}{\textbf{Model}} & \multicolumn{2}{c}{\textbf{Easy}} & \multicolumn{2}{c}{\textbf{Medium}} & \multicolumn{2}{c}{\textbf{Hard}} & \multicolumn{2}{c}{\textbf{Extra}} & \multicolumn{2}{c}{\textbf{All}}  \\ \cmidrule{2-11}
& \textbf{EX} & \textbf{EM} & \textbf{EX} & \textbf{EM} & \textbf{EX} & \textbf{EM} & \textbf{EX} & \textbf{EM} & \textbf{EX}& \textbf{EM} \\\midrule
& \multicolumn{10}{c}{\textbf{\spider}}\\ \midrule
$\text{FastRAT}_{\text{ext}}$ & $86.2$ & $81.3$ & $72.2$ & $66.0$ & $60.0$ & $51.6$ & $48.5$ & $33.9$ & $68.7$ & $60.9$\\\midrule
\multicolumn{11}{l}{\texttt{deepseek-coder-33b-instruct}}\\
\astres (w/ Graphix-T5) & $ 89.6 $ & $85.5 $ & $ 88.6 $ & $ 66.4 $ & $ 73.9 $ & $ 50.1 $ & $ 58.8 $ & $ 28.0 $ & $ 80.7 $ & $ 60.7 $ \\\midrule
\multicolumn{11}{l}{\texttt{gpt-4}}\\
\astres (w/ Graphix-T5) & $91.9$ & $87.4$ & $90.3$ & $80.4$ & $81.2$ & $66.1$ & $74.2$ & $47.6$ & $86.0$ & $73.4$ \\\midrule
& \multicolumn{10}{c}{\textbf{\cspider}}\\ \midrule
$\text{FastRAT}_{\text{ext}}$ & $-$ & $67.2$ & $-$ & $49.9$ & $-$ & $41.5$ & $-$ & $ 12.9$ & $-$ & $45.5$\\\midrule
\multicolumn{11}{l}{\texttt{deepseek-coder-33b-instruct}}\\
\astres (w/ $\text{FastRAT}_{\text{ext}}$) & $ - $ & $79.7 $ & $ - $ & $ 58.2 $ & $ - $ & $ 40.2 $ & $  $ & $ 22.4 $ & $ - $ & $ 52.9 $ \\\midrule
\multicolumn{3}{l}{\texttt{gpt-4}}\\
\astres (w/ $\text{FastRAT}_{\text{ext}}$) & $-$ & $81.2$ & $-$ & $67.7$ & $-$ & $53.2$ & $-$ & $29.9$ & $-$ & $61.1$\\
\bottomrule
\end{tabular}
\caption{\label{tab:spider_test_scores}Execution (EX) and exact match (EM) accuracy scores of our framework, on the test splits of \spider and \cspider.}
\end{table*}

\section{Schema Selection Experiments}
\label{sec:extra_schema_selection_appendix}
Table \ref{tab:extra_schema_selection} includes experiments that compare the proposed schema selection strategy against a dense retriever to capture semantic matching signals. We employ the pre-trained \texttt{all-mpnet-base-v2} model~\cite{Song2020} from SentenceTransformers~\cite{Reimers2019} to encode the semantic names of individual columns. A vector index of column embeddings is built for each database. A sub-schema is selected through column retrieval as explained in Appendix \ref{subsec:schema_val_select_appendix}, but with a dense retriever by computing the cosine similarity between the embeddings of each question and column. Approximated queries are predicted by $\text{FastRAT}_{\text{ext}}$.

Our findings remain consistent with the ones we noted in Table~\ref{tab:recall_compres}. Using a dense retriever, we fail to achieve Recall and Schema Shortening that both meet the standards of our proposed method (i.e. Approx. Query + BM25 with dynamic top-$k$). For roughly the same extent of schema shortening, our method consistently yields higher schema elements recall.

\begin{table*}[t]
    \centering
    \begin{tabular}{l c c}
    \toprule
    \textbf{Schema Selection Setup} & \textbf{Recall} & \textbf{Schema Shorten.} \\
    \midrule
    Gold Query & $100.0$ & $71.3$ \\
    \midrule
    BM25 (top-$10$) & $92.0$ & $36.5$ \\
    BM25 (top-$20$) & $98.3$ & $14.1$ \\
    Dense (top-$10$) & $78.7$ & $37.4$ \\
    Dense (top-$15$) & $86.8$ & $23.4$ \\
    Approx. Query & $86.8$ & $71.3$ \\
    \midrule
    Approx. Query + BM25 (top-$7$) & $93.3$ & $50.4$ \\
    Approx. Query + BM25 (top-$10$) & $97.0$ & $37.3$  \\
    Dense (top-$6$) + BM25 (top-$10$) & $97.1$ & $31.7$ \\
    Approx. Query + BM25 (dynamic top-$k$) & $97.2$ & $49.0$ \\
    \bottomrule

    \end{tabular}
    \caption{Recall and Schema Shortening across different schema selection setups on the development split of \spider. Approximated queries are predicted by $\text{FastRAT}_{\text{ext}}$.}
    \label{tab:extra_schema_selection}
\end{table*}

\end{document}